%% file: emnlp-ijcnlp-2019.tex
\documentclass[11pt,a4paper]{article}
\usepackage[hyperref]{emnlp-ijcnlp-2019}
\usepackage{times}
\usepackage{latexsym}
\usepackage{tabularx}
\usepackage{graphicx}
\usepackage{booktabs}
\usepackage{amsfonts}       
\usepackage{nicefrac}       
\usepackage{microtype}      
\usepackage{multirow}
\usepackage{xcolor}
\usepackage{caption}
\usepackage{amsmath}
\usepackage{float} 

\usepackage{url}

\aclfinalcopy 

\DeclareMathOperator*{\argmin}{\arg\!\min}

\title{
Entity Projection via Machine Translation for Cross-Lingual NER
}

\author{Alankar Jain \qquad Bhargavi Paranjape \qquad Zachary C. Lipton \\ 
Carnegie Mellon University \\
Pittsburgh, USA \\
{\tt \{alankarjain91,bhargavi22294\}@gmail.com}, {\tt zlipton@cmu.edu}}

\date{}

\begin{document}
\maketitle

\begin{abstract}
    \input{sections/abstract_2.tex}
\end{abstract}

\input{sections/intro_v2.tex}
\input{sections/related_work.tex}

\input{sections/method.tex}

\input{sections/experiments.tex}

\input{sections/error_analysis.tex}

\input{sections/discussion.tex}

\input{sections/conclusion.tex}

\bibliography{emnlp-ijcnlp-2019}
\bibliographystyle{acl_natbib}

\end{document}

%% file: sections/abstract_2.tex
Although over $100$ languages are supported by strong off-the-shelf machine translation systems, 
only a subset of them possess large annotated corpora for \emph{named entity recognition}.
Motivated by this fact, we leverage machine translation 
to improve annotation-projection approaches to cross-lingual named entity recognition. 
We propose a system that improves over prior entity-projection methods by:
(a) leveraging machine translation systems twice: 
first for translating sentences and subsequently for translating entities;
(b) matching entities based on orthographic and phonetic similarity; and 
(c) identifying matches based on distributional statistics derived from the dataset. 
Our approach improves upon current state-of-the-art methods
for cross-lingual named entity recognition 
on $5$ diverse languages by an average of $4.1$ points. 
Further, our method achieves state-of-the-art $F_1$ scores for 
Armenian, outperforming even a monolingual model
trained on Armenian source data.\footnote{Code for our paper can be found at: 
\url{https://github.com/alankarj/cross_lingual_ner}}

%% file: sections/intro_v2.tex
\section{Introduction}
\label{sec:intro}

While machine learning methods for various 
\emph{Natural Language Processing (NLP)} tasks 
have progressed rapidly,
the benefits
accrue disproportionately 
among languages
endowed with large annotated corpora.
Owing to the dependence of state-of-the-art deep learning approaches 
on massive amounts of data,
creating suitable datasets can be prohibitively expensive. 
This asymmetry between resource-rich and 
relatively under-resourced languages
has inspired work on \emph{cross-lingual} approaches 
that leverage annotated datasets from the former
to build strong models for the latter.

This paper focuses on cross-lingual approaches 
to \emph{Named Entity Recognition (NER)},
owing to 
NER's importance as a core component
in \emph{information retrieval} and \emph{question answering} systems. 
Specifically, we focus on \textbf{medium-resource} languages.
We define these to be languages 
for which although annotated NER corpora do not exist, 
off-the-shelf \emph{Machine Translation (MT)} systems, 
such as Google Translate%
\footnote{https://cloud.google.com/translate/}, do. 
We are motivated by the fact that although there are fewer than $50$ %
languages for which large NER datasets (greater than $200$k tokens) 
with gold annotations are publicly available%
\footnote{http://damien.nouvels.net/resourcesen/corpora.html}, 
many more languages are supported by good-quality MT  \citep{wu2016google}.
Google Translate alone supports $103$ languages%
\footnote{https://cloud.google.com/translate/docs/languages}, 
many of which have either no, or only small, NER datasets. 






We address the setting where
annotated corpora exist in the source 
(resource-rich) language---English in our experiments---but for the target (medium-resource) language, 
we can only afford to label a small validation set. 
We tackle this problem by first creating 
an unlabeled dataset in the target language 
by translating each sentence in the source dataset to the target language. 
For MT, we use Google Translate,
motivated by its large coverage. 
%
Next, we annotate this dataset 
via
\emph{entity projection}---first aligning every entity in a source sentence 
with its counterpart in the corresponding target sentence 
(\textit{entity alignment}) 
and then projecting the tags from source to target 
in the aligned entity pairs (\textit{tag projection}). 
One consequence of 
relying on MT
as opposed to word-by-word or phrase-by-phrase translation
is that the entity projection step can be difficult, 
owing to the frequency with which original sentences
and their translated counterparts are not word-for-word aligned.

\begin{figure*}[ht!]
    \centering
    \includegraphics[width=0.75\textwidth]{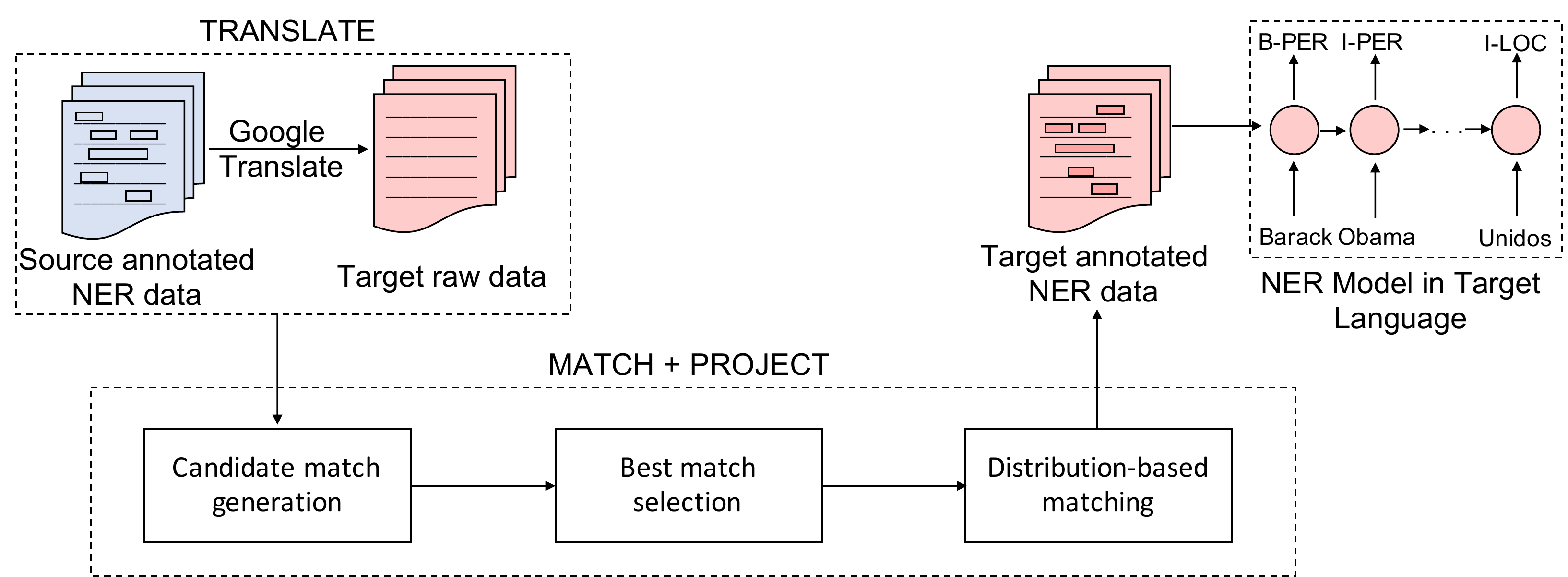}
    \caption{A schematic diagram representing the chief steps in our method.}
    \label{fig:pipeline}
\end{figure*}

Our proposed solution to this problem consists of
(a) leveraging MT again for translating entities;
(b) matching entities based on orthographic and phonetic similarity;
and (c) identifying matches based on distributional statistics derived from the dataset. 
\emph{Importantly, while our method depends on several matching heuristics,
these techniques are remarkably portable across target languages,
requiring the tuning of only two hyperparameters.}
Our method achieves state-of-the art $F_1$ scores for cross-lingual NER 
for Spanish ($+1.1$ points), German ($+1.4$ points), and Chinese ($+5$ points) 
and beats state-of-the-art baselines on Hindi ($+2.1$ points) and Tamil ($+5$ points). 
Further, it achieves state-of-the-art $F_1$ scores for Armenian, 
a medium-resource language, 
beating a monolingual model trained on Armenian source data by $0.4$ points.

%% file: sections/related_work.tex
\section{Related Work}
\label{sec:related-work}

Cross-lingual approaches have been applied to 
many NLP tasks, including part-of-speech tagging 
\citep{yarowsky2001inducing, xi2005backoff, das2011unsupervised, tackstrom2013token}, 
parsing \citep{hwa2005bootstrapping, zeman2008cross, smith2009parser, ganchev2009dependency}, 
and semantic role labeling 
\citep{tonelli2008frame, pado2009cross, kozhevnikov2013cross, kozhevnikov2014cross}. 
Prior cross-lingual NLP papers cleave roughly into 
two distinct approaches: 
\emph{direct model transfer} and \emph{annotation projection}. 

\subsection{Direct model transfer}
These approaches apply models trained on the source language 
absent modification (to the model) to data from the target language
by exploiting a shared representation for the two languages 
\citep{tackstrom2012cross, bharadwaj2016phonologically, chaudhary2018adapting, kozhevnikov2014cross, ni2017weakly}.
%
However, direct model transfer techniques 
face a problem when applied to markedly dissimilar languages: 
they lack of lexicalized (especially character-based) features, 
which are known to have predictive power for tasks such as NER. 
\citet{xie2018neural} provide evidence 
for this in the cross-lingual setting,
comparing otherwise similar annotation projection approaches
that differ in their use of lexicalized features.


\subsection{Annotation projection}
These approaches to cross-lingual NLP 
train a model in the target language.
This requires first projecting annotations 
from the source data to the (unlabeled) target data. 
Many approaches in this category rely upon parallel corpora 
\cite{yarowsky2001inducing, hwa2005bootstrapping, zeman2008cross, ehrmann2011building, fu2011generating, ni2017weakly},
first annotating the source data using a trained model 
and then projecting the annotations. 
Only a few works explore the use of MT 
to first translate a gold annotated corpus 
to obtain a \emph{synthetic} parallel corpus 
and then project annotations \citep{tiedemann2014treebank}. 
\citet{shah2010synergy} go in the opposite direction, 
translating target to source using Google Translate, 
annotating the translated source sentences 
using a trained NER system and then projecting annotations back.

When projecting annotations, one encounters the problem of word alignment. 
Most of the existing works \citep{yarowsky2001inducing, shah2010synergy, ni2017weakly} 
rely upon unsupervised alignment models from statistical MT literature, 
such as IBM Models 1-6 \cite{brown1993mathematics, och2003systematic}. 
Other works focus on low-resource settings \citep{mayhew2017cheap, xie2018neural} 
perform translation word-by-word or phrase-by-phrase, 
and thus do not need to perform word alignment. 
Several papers explore heuristics such as using Wikipedia links across languages 
to align entities \citep{richman2008mining, nothman2013learning, al2015polyglot}, 
matching tokens based on their surface forms 
and transliterations either in an unsupervised manner \cite{samy2005proposal, ehrmann2011building}
or as features in a supervised model trained 
on a small \emph{seed} dataset \cite{feng2004new}. 
Many of these papers often rely on language-specific features \cite{feng2004new}
and evaluate their alignment methods on only a few languages.

To our knowledge, few works effectively use translation
for annotation projection for NER, 
especially for medium-resource languages 
for which strong MT systems exist.
Motivated by this research gap, we explore the use of MT systems 
for translating the dataset and for annotation projection 
and thus do not rely on parallel corpora. 
However, we demonstrate the efficacy of our projection method in all three settings:
(a) translation from source to target,
(b) using parallel corpora and 
(c) translation from target to source.

%% file: sections/method.tex
\section{The Translate-Match-Project Method \label{sec:tmp}}

\begin{figure}[t!]
    \centering
    \includegraphics[width=0.45\textwidth]{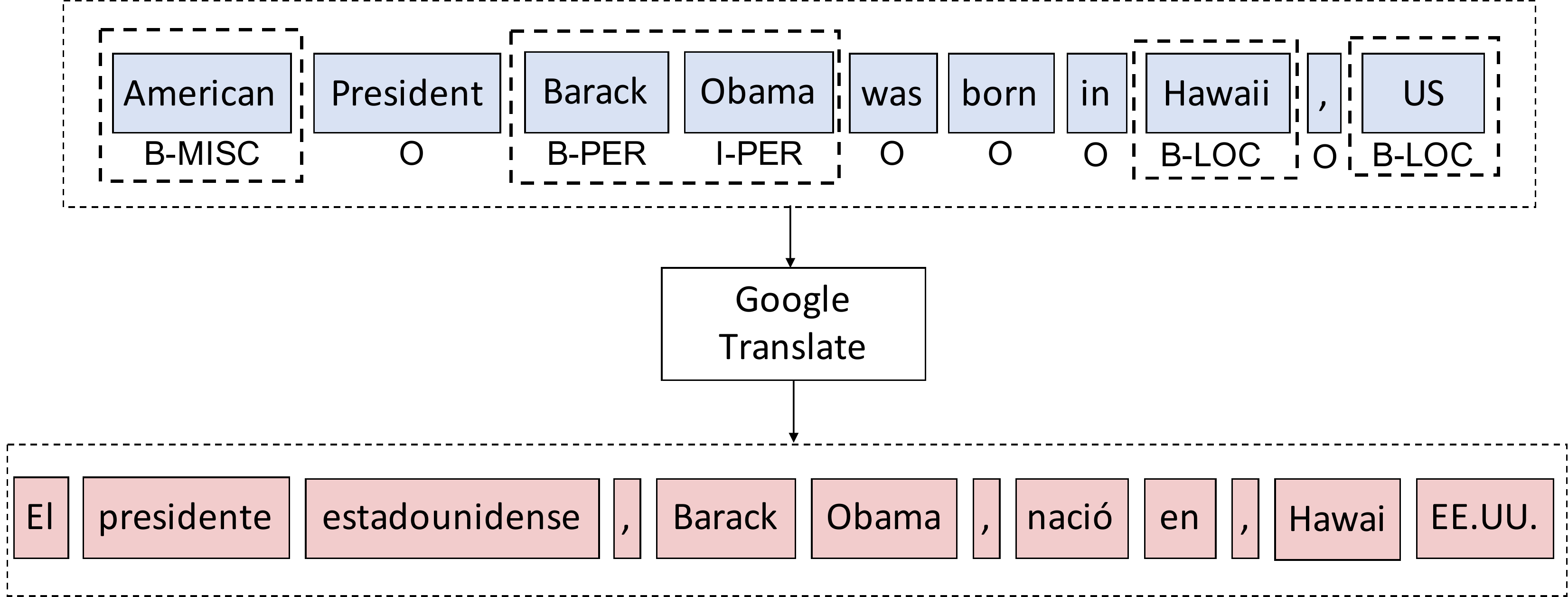}
    \caption{The \emph{Translate} step of our method.}
    \label{fig:translate}
\end{figure}

In our formulation, we are given an annotated NER corpus in the source language: 
$\mathcal{D}^S_A = \{(x^{Si}, y^{Si}): i = 1, 2, ..., N\}$, 
where $x^{Si} = (x^{Si}_1,...,x^{Si}_{L^{Si}})$ 
is the $i$th source sentence with $L^{Si}$ tokens 
and $y^{Si} = (y^{Si}_1,...,y^{Si}_{L^{Si}})$ 
are the NER tags from a fixed tag set. 
We work with four tags: PER (person), ORG (organisation), 
LOC (location), and MISC (miscellaneous). 
We follow the commonly-used IOB (Inside Outside Beginning) 
tagging format \cite{ramshaw1999text}. 
In our experiments, we work with English as the source language
due to the availability of high-quality annotated corpora, 
e.g., CoNLL 2002 \citep{sang2002introduction} 
and OntoNotes 4.0 \citep{weischedel2011ontonotes}. However, our method can easily be applied to any other resource-rich source language as well.
See Figure \ref{fig:translate} for an example 
of an annotated source sentence (blue), $(x^{Si}, y^{Si})$.

We are given a small labeled development set data 
(but no training data)
in the target language $T$ for tuning hyperparameters.
Our method, denoted \emph{Translate-Match-Project (TMP)}, proceeds in three steps:
%
First, we translate the annotated corpus in $S$ to $T$
using an off-the-shelf MT system (Google Translate). 
This results in an un-labeled dataset in the target language, 
$\mathcal{D}^T = \{x^{Ti}: i = 1, 2, ..., N\}$ 
(Figure \ref{fig:translate});
%
Second, we identify and tag all named entities in the translated target sentences 
by entity projection, which involves entity alignment and tag projection.
We perform entity alignment by first constructing 
a set of potential matches in the target sentence 
for every entity in the source sentence 
(\textit{candidate match generation}, Section \ref{sec:cand-match-gen}) 
and then by selecting the best matching pairs of source and target entities 
(\textit{best match selection}, Section \ref{sec:best-match-select}); 
Third, after alignment, we project the tag type (PER, LOC, etc.) from the source to the target entity 
in every pair of aligned entities 
by adhering to the IOB tagging scheme in target. 
In Figure \ref{fig:pipeline}, we depict the complete pipeline.




\subsection{Candidate match generation \label{sec:cand-match-gen}}
To generate candidate matches for an entity in a source sentence, 
we construct a set of its potential translations 
and then find matches for each  
in the corresponding target sentence. 
We find these matches by token-level matching and then
\emph{concatenate matched tokens to obtain multi-token matches}. We drop the index $i$ below for ease of notation.

\paragraph{Token-level matching \label{sec:tok-level}}
Consider a source entity $e^S=(x^S_{j},...,x^S_{k})$. 
For every $e^S \in \mathcal{E}^S$, 
where $\mathcal{E}^S$ is the set of all entities in a source sentence, 
we obtain a set of potential translations in the target language, $\mathcal{T}(e^S)$, 
via MT.
However, in some cases, 
translating a standalone entity
produces a different translation 
from that which emerges when translating 
a full sentence.
For example, Google Translate maps the source entity ``UAE"
to ``Emiratos \'Arabes Unidos'' in most sentences 
but the word-by-word translation is ``EAU''. 
Similarly, the (person) name ``Tang'' (e.g., ``Mr. Tang") 
remains ``Tang'' in translated sentences, 
but is translated to ``Espiga'' 
(Spanish for ``spike'', synonymous 
with the English word ``tang''). 
We address these problems by augmenting $\mathcal{T}(\cdot)$ 
with translations
from \emph{publicly available} bilingual lexicons 
(``UAE" translates to ``Emiratos \'Arabes Unidos" in one of the lexicons we use) 
and retain
a copy of the source entity 
(``Tang" will now find a match in the target sentence). 
Finally, $\mathcal{T}($``UAE"$)$ looks roughly like: 
\{``EAU" [Google Translate], ``UAE" [copy], ``Emiratos \'Arabes Unidos" [lexicon]\}. 
We note that lexicons exist for a large number of languages today\footnote{Panlex \cite{kamholz2014panlex} has lexicons for 10k different languages.}.
However, we demonstrate that our method 
also works in absence of such lexicons 
in our case study for Armenian 
(Section \ref{subsec:armenian}).


Next, we tokenize each candidate translation in $\mathcal{T}(e^S)$ 
to obtain a set of translation tokens for $e^S$, $\mathcal{T}^{w}(e^S)$. 
For example, $\mathcal{T}^{w}$(``UAE") = \{``EAU", ``UAE", ``Emiratos", ``\'Arabes", ``Unidos"\}. 
We do this to allow for soft token-level matches 
because we observed empirically 
that matching exact entity phrases might result in few matches.
Next, we obtain a match for each hypothesis token $h \in \mathcal{T}^{w}(e^S)$
by matching it with each reference token $x^S_{l} \; \forall \; l \in \{1,...,L^T\}$ 
in the target sentence $x^T = (x^T_{1},...,x^T_{L^T})$ of length $L^T$.
This match is carried out at
(a) the orthographic (surface form) level; 
and (b) the phonetic level, 
by matching transliterations in the International Phonetic Alphabet (IPA) of the two tokens. 
In either case, we look for the longest sequence of characters in $h$
that are an affix (prefix or suffix) of $x^T_{l}$. 
This soft \emph{affix-matching heuristic} allows 
for inflection in morphologically-rich target languages. 
The (token-level) score for the match is given as follows:
\begin{align*}
    s^{w}(h, x^T_{l}) &= \min \Big\{\frac{n_{l}}{L_{h}}, \frac{n_{l}}{L_{x^T_{l}}}\Big\}
\end{align*}
Here, $L_{h}$ and $L_{x^T_{l}}$ are the lengths (in characters)
of the hypothesis and reference tokens
and $n_{l}$ is the number of matching characters. 
We take minimum in order to enforce a stricter notion of fractional (soft) match. 
For example, the phrase ``German first-time registrations...'' [English]
gets translated to ``Los registros Alemanes por primera..." [Spanish]. 
Using our matching heuristic, ``Alem\'an'' $\in \mathcal{T}^{w}$(``German") 
matches to the reference token ``Alemanes'' with a score of $0.5$, 
since $n_{l} = 4$ (``Alem'') and $L_{x^T_{l}} = 8 > L_{h} = 6$.
Next, we define the matching (entity-level) score between a source entity $e^S$ 
and any target token $x^T_{l}$ as follows:
\begin{align*}
s^{e}(e^S, x^T_{l}) &= \max_{h \in \mathcal{T}^{w}(e^S)} s^{w}(h, x^T_{l})
\end{align*}

\begin{figure}[t!]
    \centering
    \includegraphics[width=0.45\textwidth]{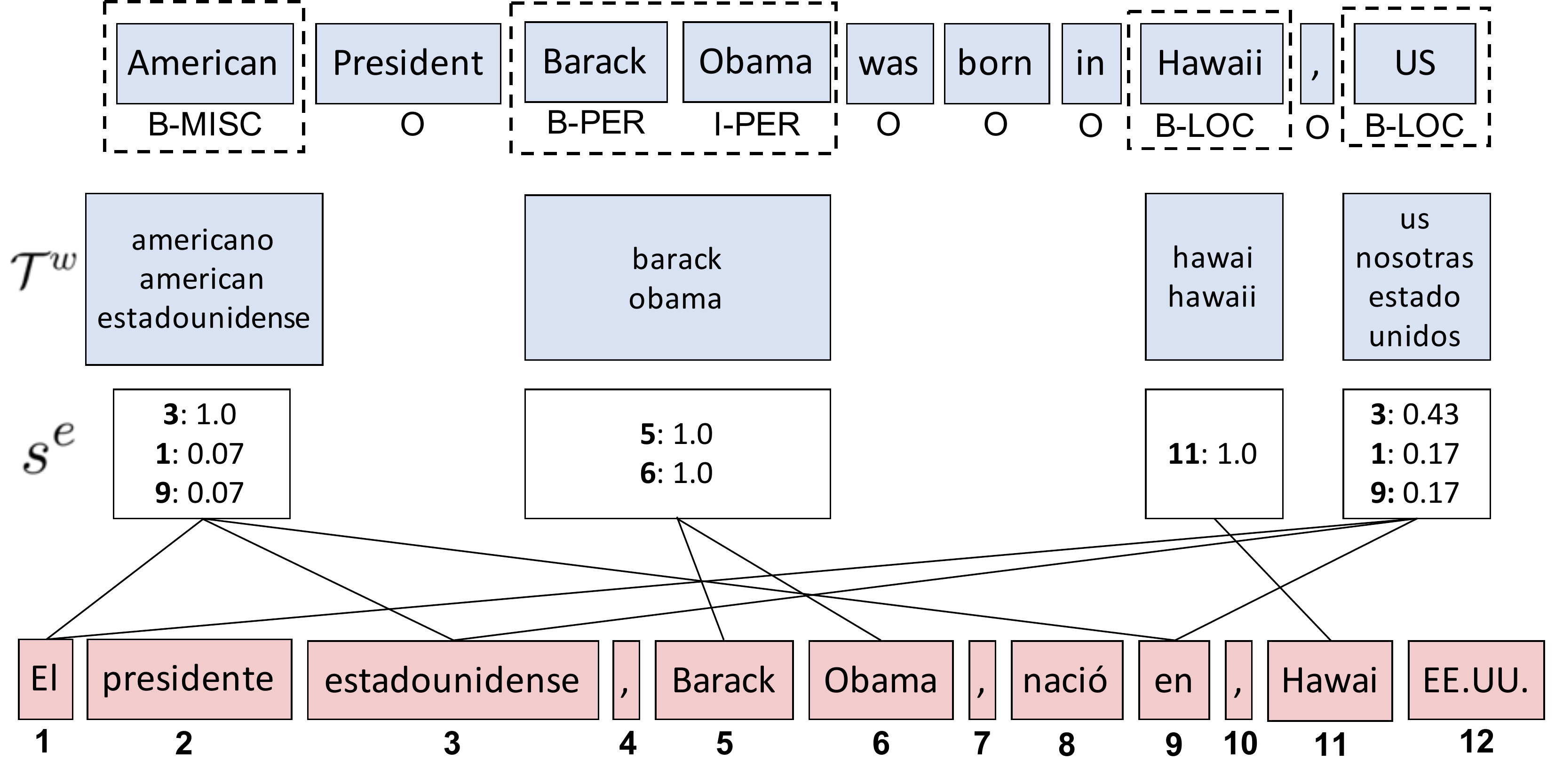}
    \caption{Token-level matching: Blue boxes in row-2 ($\mathcal{T}^w$) show sets of potential translations and white boxes in row-3 ($s^e$) show target tokens (numbered) each source entity can match with, with their scores.}
    \label{fig:tok-level}
\end{figure}

Note that the token-level scores, $s^w$, include scores based on both orthographic and phonetic match, and thus, the entity-level scores, $s^e$, correspond to the best token-level match (orthographic or phonetic) between any hypothesis token $h \in \mathcal{T}^{w}(e^S)$ and a target token $x^T_l$.
In Figure \ref{fig:tok-level}, we depict the token-level matching procedure. 
The score between the entity ``American''
and the target ``estadounidense"(labeled 3 in the figure) 
is the maximum over matching scores between 
any token in $\{$``americano", ``american'',  ``estadounidense''$\}$ 
and the target token (``estadounidense"), i.e., $1.0$ (exact match) 
since the scores for the first two tokens in $\mathcal{T}^w(\cdot)$ are $0$. 
Note some artifacts of token-level matching: (a) our matching heuristic currently handles prefixes or suffixes, but can potentially be extended with character edit distance for other types of affixes (e.g., circumfix)
(b) every target token can match with multiple source entities 
(for e.g., ``El'' matches with both ``American" and ``US") 
and (c) some source entities might fail to find their true match 
(``US" fails to match with ``EE.UU.'', 
a possibly erroneous translation of ``US" provided by Google Translate). 
Further, many matches are of very poor quality 
(especially those with stop words such as ``El" and ``en''). 
We address these issues and describe how to convert these token-level matches 
into spans to get multi-token target entities next.

\paragraph{Span match generation \label{sec:span-match}}
After token-level matching, we construct a list 
of potential entity spans in the target sentence 
that match with a given source entity $e^S$ 
by grouping adjacent target tokens for which a token-level matching score $s^e(e^S, \cdot)$ is above a threshold $\delta$ (to remove spurious matches).
In other words, we construct the following set:
\begin{align*}
    \mathcal{M}(e^S) = \{\text{span}(q,r) : s^{e}(e^S, x^T_{u}) \geq \delta \\
    \; \forall \; q \leq u \leq r \}
\end{align*}
Here, $\text{span}(q,r) = (x^T_{q},...,x^T_{r})$ 
is the phrase spanning tokens indexed from $q$ to $r$ 
($1 \leq q, r \leq L^T$) in the target sentence.
Further, we require $\text{span}(q, r)$ to be maximal 
in the sense that $\forall \; q'<q$ and $\forall \; r'>r$, $\text{span}(q',r') \notin \mathcal{M}(\cdot)$, i.e., 
any target token before or after the span 
have at best a weak match ($s^{e}(e^S, \cdot) < \delta$) with $e^S$.

\begin{figure}[htb]
    \centering
    \includegraphics[width=0.45\textwidth]{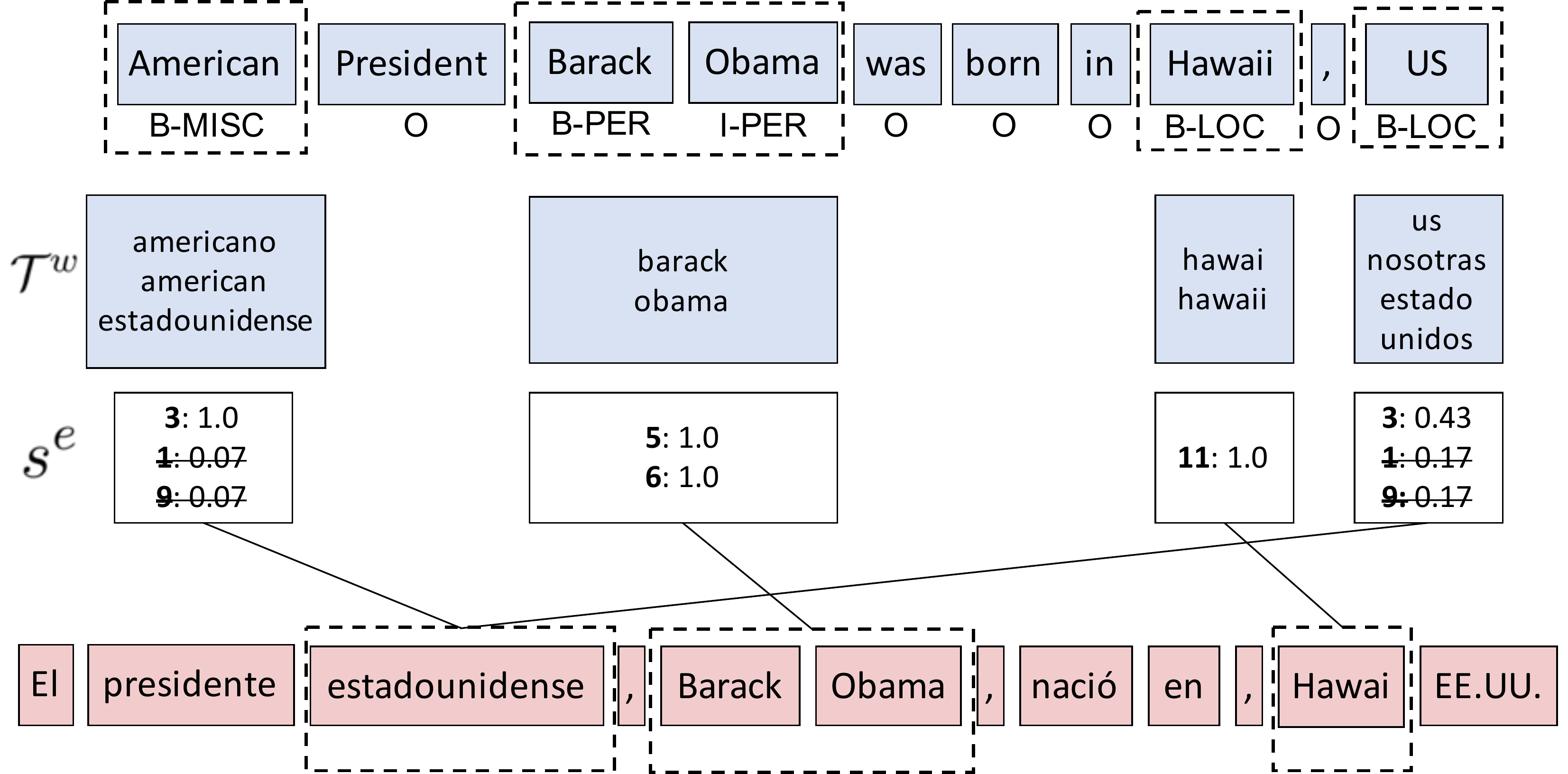}
    \caption{Span match generation: Adjacent target tokens with matching scores higher than a threshold (0.25 here) are concatenated to form span-level matches.}
    \label{fig:span-level}
\end{figure}


In our running example, choosing $\delta = 0.25$ 
results in the spans shown in Figure \ref{fig:span-level}, 
eliminating spurious matches (with ``El" and ``en") 
and concatenating ``Barack" and ``Obama" in the target sentence. 
However, the token ``estadounidense" is still matched 
with two different source entities. 
We solve this problem in the next step.


\subsection{Best match selection \label{sec:best-match-select}}
For selecting the best matching pair of entities, 
we first expand the set of potential translations $\mathcal{T}(\cdot)$
to include all possible token-level permutations of the translations. 
We call this set $\mathcal{T}^{p}(\cdot)$. For example, $\mathcal{T}^{p}$(``UAE'') = \{``EAU'', ``UAE'', ``Emiratos \'Arabes Unidos'', ``Emiratos Unidos \'Arabes'', ``\'Arabes Emiratos Unidos'', ...\}.
Then, we greedily align $e^S$ with the target entity span from the set $\mathcal{M}(e^S)$, 
with the least character edit distance $d_E(\cdot, \cdot)$ 
from any translation in $T^{p}(e^S)$, i.e.,
\begin{align*}
e^T = \argmin_{\text{span}(\cdot, \cdot) \in \mathcal{M} (e^S)} d_{E}(e^S, \text{span}(\cdot, \cdot))
\end{align*}
In this manner, we form 
aligned entity pairs $(e^S, e^T)$, 
along which tags can then be projected. 
In our running example, since the edit distance between ``estadounidense'' 
(in $\mathcal{T}^p(\cdot)$) and the target single-token span ``estadounidense'' is $0$, 
and is lower than that with ``estado'', 
we match ``estadounidense" with ``American'', tagging it B-MISC.






\subsection{Distribution-based matching}
After selecting best matching pairs, 
there still remain some source entities 
that do not find any matching target entity (``US'' in our example). 
These arise either due to significant differences between word-by-word 
and contextual sentence-level translations either due to literal
(e.g., ``West Bank'' gets translated to ``Cisjordania''
in sentences and ``Banco Oeste'' otherwise) 
or possibly incorrect translations
(e.g., ``U.S." gets translated to  ``EE.UU.'' in sentences and ``NOSOTRAS'' otherwise). 

\begin{figure}[htb]
    \centering
    \includegraphics[width=0.45\textwidth]{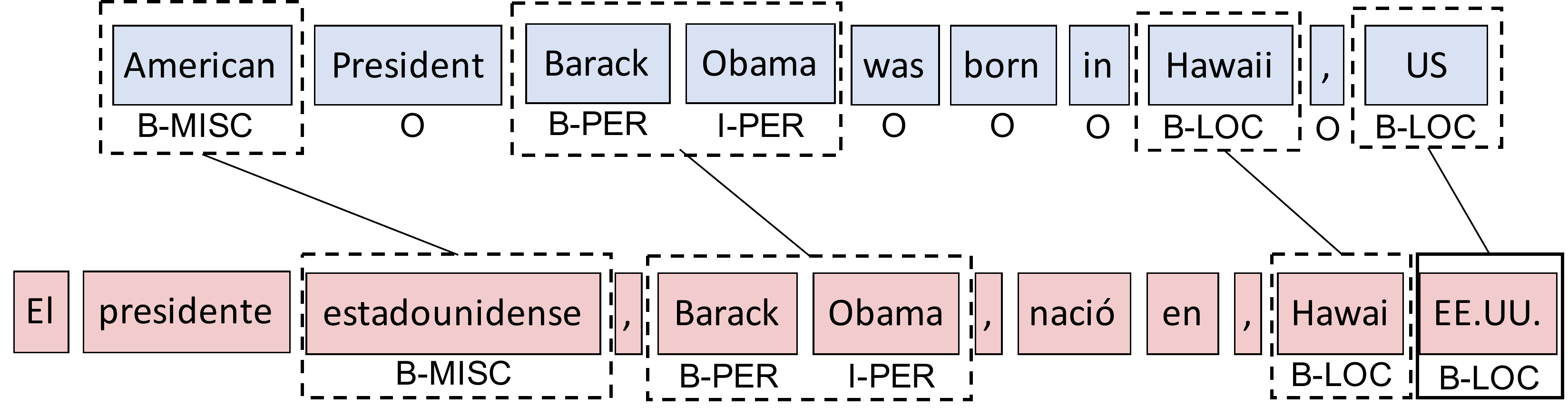}
    \caption{Final aligned pairs and projected tags.}
    \label{fig:best-match}
\end{figure}

We remedy this by exploiting corpus-level 
consistency in such discrepancies.
For every unmatched source entity, 
we construct a set of top-$k$ potential matches 
ordered by their tf-idf (term frequency---inverse document frequency) scores, 
where tf is calculated over all sentences 
containing at least one unmatched entity 
and the idf score is calculated over the entire dataset 
to severely penalize commonly occurring tokens. 
Finally, we match each unmatched source entity 
with an unmatched span in its top-$k$ list with the highest tf-idf score. Figure \ref{fig:best-match} shows the final matches and tags.


\begin{table*}[!h]
\small
\centering
\begin{tabularx}{0.95\textwidth}{cccccccc}
\toprule
\bf Method & \bf Spanish & \bf German & \bf Dutch & \bf Chinese & \bf Hindi & \bf Tamil & \bf Average \\
\toprule
TMP & \bf 73.5 $\pm$ 0.4 & \bf 61.5 $\pm$ 0.4 & 69.9 $\pm$ 0.4 &
\bf 50.1 $\pm$  0.2 &
\bf 41.7 $\pm$  1.3 & \bf 33.8 $\pm$ 2.2 & \bf 55.1 $\pm$ 0.8 \\
\midrule
fast-align & 65.0 $\pm$ 1.2 & 
60.1 $\pm$ 0.9 &
67.6 $\pm$ 0.7 &
45.1 $\pm$ 0.8 &
39.6 $\pm$ 1.1 & 
28.8 $\pm$ 1.8 & 
51.0 $\pm$ 1.1 \\
BWET & 72.4 $\pm$ 0.6 & 
57.8 $\pm$ 0.1 & \bf 70.4 $\pm$ 1.2 &
3.51 $\pm$ 0.8 &
26.6 $\pm$ 0.8 & 
15.6 $\pm$ 0.9 & 
48.5 $\pm$ 0.7 \\
Co-decoding & 65.1 & 58.5 & 65.4 & - & - & - & - \\
Polyglot-NER & 63.0 & - & 59.6 & - & - & - & - \\
\midrule
Monolingual & 86.3 $\pm$ 0.4 & 78.2 $\pm$ 0.4 & 86.4 $\pm$ 0.2 & 
68.59 $\pm$ 0.3 & 
65.8 $\pm$ 1.2 & 51.8 $\pm$ 1.0 & 73.7 $\pm$ 0.6 \\
\bottomrule
\end{tabularx}
\caption{Test $F_1$ scores for our method (TMP), 4 cross-lingual baselines and a model trained on monolingual data.}
\label{tab:results}
\end{table*}

%% file: sections/experiments.tex
\section{Experimental Evaluation}
\paragraph{Data}
In order to compare our method against benchmarks reported for prior approaches, 
we evaluate its performance on three European languages: 
Spanish (\textit{es}), Dutch (\textit{nl}) and German (\textit{de}). 
Further, for a more extensive evaluation, 
we conduct additional experiments in
an Indo-Aryan language (Hindi (\textit{hi})),
a Dravidian language (Tamil (\textit{ta})), 
and Simplified Chinese (\textit{zh}).

For all languages except Chinese, 
we use English NER training data 
from the CoNLL 2003 shared task \cite{sang2003introduction} 
to translate into the target language. 
For Chinese, we sample the same number of sentences as
in the CoNLL 2003 corpus (14,041) from the OntoNotes 4.0 (2012) 
dataset for English \cite{weischedel2011ontonotes} to minimize distribution shift from Chinese development data. 
The development and test datasets for Spanish, Dutch and German 
are obtained from CoNLL 2002 \cite{sang2002introduction} 
and CoNLL 2003 shared tasks. 
For Hindi and Tamil, we obtain the NER corpus from FIRE 2013 shared task%
\footnote{http://au-kbc.org/nlp/NER-FIRE2013/}. 
Since this task doesn't provide the test dataset, 
we create our own splits: two-thirds for training 
and one-sixth each for development 
and test (to match with the proportions in the CoNLL dataset). 
For Chinese, we use the OntoNotes 4.0 development and test datasets. 
English, Spanish, Dutch and German contain PER, ORG, LOC and MISC tags, 
while Hindi, Tamil and Chinese were preprocessed 
to contain only PER, LOC and ORG tags.
We use MUSE ground-truth bilingual lexicons%
\footnote{https://github.com/facebookresearch/MUSE} (gold lexicon) for augmenting the set of potential entity translations and use Epitran \cite{mortensen2018epitran} for obtaining IPA transliterations.


\paragraph{Baselines}
We compare against
four other annotation projection approaches 
that have achieved state-of-the-art results on some of our datasets. 
\citet{xie2018neural} (BWET) use a bilingual lexicon 
induced using monolingual corpora \cite{conneau2017word} 
to translate each source sentence word-by-word 
and then copy the corresponding NER tags using gold lexicons.
As a ceiling for their method, \citet{mayhew2017cheap} used Google Translate 
with fast-align \cite{dyer2013fastalign} (fast-align), 
an unsupervised expectation maximization based algorithm, for entity alignment. 
Since this algorithm can produce multiple matches for a given source entity, 
we post-process the alignments produced by this algorithm 
and select the longest match and then project tags 
in the same way as our method.
Our third baseline is \citet{ni2017weakly} (Co-decoding), 
who use a co-decoding scheme on two different NER models.
We also compare our method with Polyglot-NER \cite{al2015polyglot} 
who use Wikipedia links to project entities. 
Finally, we also compare our performance with a model trained on annotated data in target language (Monolingual).



\paragraph{NER Model}
We use the state-of-the-art neural NER tagging model from \cite{xie2018neural} 
to train TMP and fast-align baseline for all languages. 
This model adds a self-attention layer to the 
character- and word-based BiLSTM + CRF model due to \citet{lample2016neural}. 
For each experiment, we run our models $5$ times using different seeds 
and report the mean and standard deviation 
(as recommended by \citet{reimers2017reporting}) of $F_1$ measure.



\paragraph{Hyperparameters}
For the fast-align baseline, we tune their $\lambda$ parameter,
which controls how much the model deviates 
from perfectly diagonal alignments, for each language separately. 
For TMP, we tune $\delta$, the score threshold and $k$, the number of top candidates selected in distribution-based matching.  
We use the same hyperparameters for the NER model as \citet{xie2018neural} 
for all our experiments.



\input{sections/results.tex}
\input{sections/analysis.tex}

\input{sections/case_study.tex}

%% file: sections/results.tex
\subsection{Results}
\label{sec:results}
\begin{figure*}
    \centering
    \includegraphics[scale=0.40]{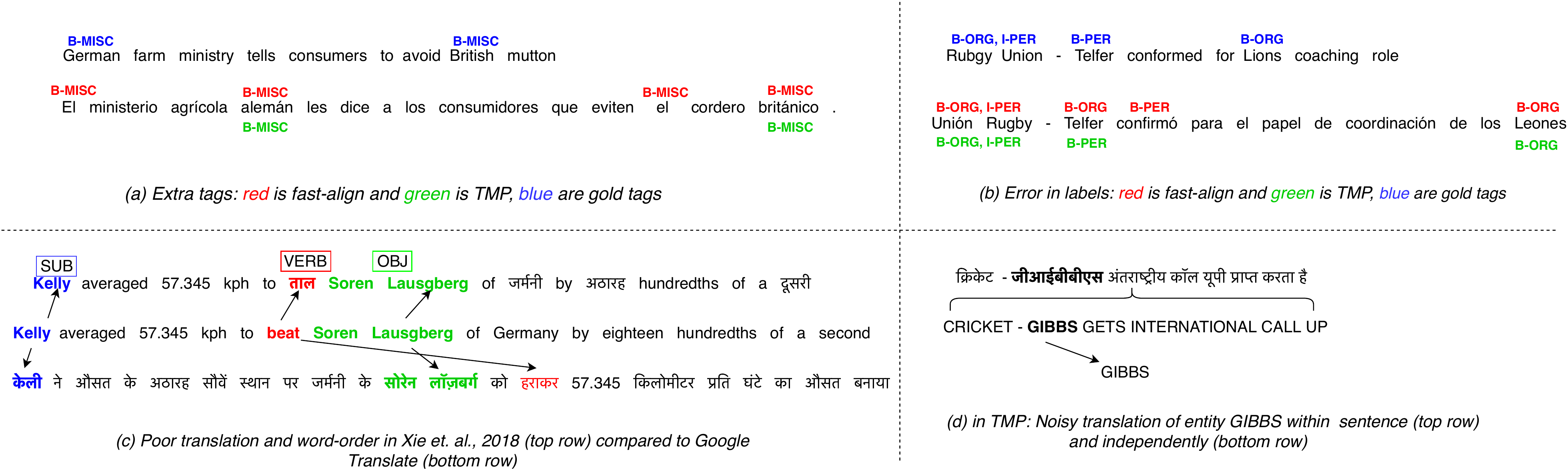}
    \caption{Examples of different errors (details in individual captions).}
    \label{fig:error_analysis}
\end{figure*}

Our technique outperforms previous state-of-the-art cross-lingual methods 
on Spanish, German, Chinese, Hindi and Tamil 
and performs competitively on Dutch (Table \ref{tab:results}). 
In particular, our method shows marked improvements over BWET, 
a word-by-word translation baseline, for languages 
such as German, Hindi, Tamil and Chinese 
that differ markedly in word ordering (with respect to English), 
demonstrating the impact of improved machine translation quality 
on final NER tagging accuracy. 
For more distant languages, word ordering can drastically affect the position of entities in a sentence,
which can hurt performance on a test set in the target language.
For instance, consider the Hindi word-by-word translation in Figure \ref{fig:error_analysis} (c), 
which is incoherent and violates the Subj-Obj-Verb ordering of Hindi. On languages that are closer to English, like Spanish and Dutch, 
the gains are comparatively modest,
indicating that word order and quality MT is not critical for such languages.

We also show improvements over the fast-align baseline, which performs unsupervised 
word-level alignment over the full sequence.
This can lead to alignment errors for named entities, 
which tend to be low-frequency words. 
Moreover, since fast-align allows 
for  multiple target words to be aligned to a given source word, 
several noisy tags are added to the target sentence 
(see Figures \ref{fig:error_analysis} (a) and (b)). 

%% file: sections/analysis.tex
\subsection{Comparison of projection settings} 
Having established the performance of TMP 
as a method for cross-lingual NER,
in this section, we conduct deeper experiments 
to evaluate the effectiveness of the matching (M) and projection (P) steps of TMP over the other projection baseline, fast-align. 
As mentioned in Section \ref{sec:related-work}, 
there are variants of the annotation projection paradigm for cross-lingual NER 
that require an entity projection step, namely 
(i) reversing the direction of machine translation and 
(ii) using parallel corpora. 
We compare MP with fast-align for Spanish and Hindi languages under both these settings.

\begin{table}[htb]
\small
\centering
\begin{tabularx}{220pt}{ccccc}
\toprule
\bf Lang. & \bf Method & \bf Forward & \bf Reverse & \bf Parallel \\
\midrule
\multirow{2}{*}{\bf es} & MP & \bf 73.5 $\pm$ 0.4 & 65.3 & 61.2 $\pm$ 1.2 \\
 & fast-align & 65.0 $\pm$ 1.2 & 57.8 & 39.3 $\pm$ 0.5 \\
\midrule
\multirow{2}{*}{\bf hi} & MP & 41.7 $\pm$ 1.3 & 47.7 & \bf 52.8 $\pm$ 1.4 \\
& fast-align & 39.6 $\pm$ 1.1 & 34.3 & 51.8 $\pm$ 1.5 \\
\bottomrule
\end{tabularx}
\caption{Performance of MP and fast-align on Forward, Reverse and Parallel settings in terms of $F_1$.}
\label{tab:analysis_paradigms}
\end{table}


\paragraph{Reversing the direction of translation} 
In this setting, we translate the target test set 
into the source language using Google Translate 
and then use the NER tagger with state-of-the-art results \textit{Flair}%
\footnote{https://github.com/zalandoresearch/flair} 
to tag entities in the translated English sentences.
Finally, we employ MP/fast-align to project the tagged entities back to the target sentence. 
As shown in table \ref{tab:analysis_paradigms}, 
MP outperforms fast-align for both Spanish and Hindi 
and performs better than the forward direction translation for Hindi. 
This can be attributed to 
(a) the inherent difficulty of NER tagging in Hindi,
which is morphologically richer than English and 
(b) the superior quality of the English NER model.

\paragraph{Parallel corpora} 
In order to remove translation errors while evaluating TMP and fast-align, 
we experiment with parallel corpora. 
For English-Spanish, we use the Europarl corpus \cite{koehn2005europarl} 
and for English-Hindi, the IIT Bombay parallel corpus \cite{kunchukuttan2017iit}. 
We again use \textit{Flair} to obtain NER tags in English, 
which are then projected to their corresponding target sentences to generate a training dataset, which is used to train an NER model in the target language. 
To minimize confounding variables, 
we sample $14$k (same as CoNLL) high quality tagged sentences 
(average confidence score $>0.9$). 
Results in Table \ref{tab:analysis_paradigms} show 
that MP once again outperforms fast-align. 
Further, it performs better than Forward for Hindi 
by a significant margin possibly because the chosen parallel corpus 
is closer in time period to the test set, thereby reducing distribution shift.



%% file: sections/case_study.tex
\subsection{Case study: Armenian \label{subsec:armenian}}
So far, we have only evaluated the performance of our method on languages 
for which large or moderately-sized gold annotated corpora already exist that provide an upper-bound for cross-lingual NER methods. Here, we evaluate our method on a true medium-resource language, Armenian. 
Recently, \citet{ghukasyan2018pioner} introduced a ground truth test corpus for Armenian
along with a train corpus with silver annotations extracted from Wikipedia. 
This test dataset is comprised of $2566$ sentences ($53$k tokens) 
from political, sports, local and world news 
between August 2012 and July 2018. 
Since the English CoNLL 2003 dataset contains sentences nearly two decades older, 
we expect to see significant distribution shift 
if we follow TMP (Forward approach). 
Further, we are not aware of any large English-Armenian parallel corpora. 
So, we choose the Reverse paradigm for this problem.
We achieve an $F_1$ score of $62.6$, which is significantly higher 
than that achieved by fast-align ($44.8$). 
Further, this is $0.4$ points higher than the current state-of-the-art model 
trained on over $160$k tokens of Armenian. Note that our model does not make use of any external resources for Armenian (gold lexicons, Epitran, etc.) other than an MT system. This provides evidence towards our proposed approach being an effective and generalizable cross-lingual NER method that can be used for rapid deployment to new languages.

%% file: sections/error_analysis.tex
\section{Analysis}

\paragraph{Measuring alignment accuracy}
Since we do not possess ground truth word alignments 
for the ``synthetic'' parallel corpus generated through translation, 
we rely on heuristics to measure the accuracy of alignments. 
We measure the annotation \emph{miss rate} among target sentences
with equal or fewer tagged entities as compared to source.
We also calculate the \emph{excess rate}, 
representing the fraction of excess entities 
among sentences with more tagged entities. 
Both methods perform similarly in terms of 
miss rate, $0.79$ \% (MP) vs $0.83$ \% (fast-align) 
on Spanish and $3.96$ \% (MP) vs $3.48$ \% (fast-align) on Hindi. 
However, fast-align seems 
to add 
more noisy annotations as compared to MP, 
with higher excess rates for both Spanish ($8.29$ \% vs $0.49$ \%) 
and Hindi ($6.35$ \% vs $2.20$ \%). 
A representative illustration of these noisy tags 
is shown in Figure \ref{fig:error_analysis} (a).
where fast-align tags frequent words like ``El", ``de", ``en" as entities.
To offer a more fine-grained evaluation of alignment performance, 
we manually annotate $100$ examples from 
the translated Spanish and Hindi training data 
and calculate precision, recall and $F_1$ score. 
MP outperforms fast-align for both the languages (Table \ref{tab:align-perf}).

\begin{table}[htb]
\small
\centering
\begin{tabularx}{220pt}{ccccc}
\toprule
\bf Lang. & \bf Method & \bf Precision & \bf Recall & \bf $F_1$ \\
\midrule
\multirow{2}{*}{\bf es} & MP & \bf 96.2 & \bf 96.7 & \bf 96.4 \\
 & fast-align & 84.6 & 87.4 & 85.9 \\
\midrule
\multirow{2}{*}{\bf hi} & MP & \bf 87.4 & \bf 77.6 & \bf 82.2 \\
& fast-align & 82.0 & 76.8 & 79.3 \\
\bottomrule
\end{tabularx}
\caption{Alignment performance on 100 sentences.}
\label{tab:align-perf}
\end{table}


\paragraph{Ablation of features for alignment}
We also conduct an ablation study (Table \ref{tab:ablation}) 
to understand the sources of our gains beyond a base model 
that uses translations only from Google Translate and orthographic affix matching. 
To this base model, we successively add various features of our method: 
phonetic matching, exact copy translations, gold lexicons
and finally distribution-based alignment (dist) of remaining entities. 
For both languages, we observe that every additional feature 
improves the performance of tagging, with the most important features 
being phonetic matching for Spanish and use of gold lexicons for Hindi. 
Interestingly, addition of phonetic matching hurts Hindi because of the low value of the threshold ($\delta=0.25$), which results in spurious matches due to phonetic matching. 
In Table \ref{tab:ablation}, we also see that the number of entities tagged
(as a fraction of total entities) increase with the introduction of almost every feature 
(however, all matches might not be correct). 
This underscores the correlation between quality of entity alignment 
and performance on the downstream tagging task. 

\begin{table}[htb]
\small
\centering
\begin{tabularx}{220pt}{ccccc}
\toprule
\bf Model & \multicolumn{2}{c}{\bf es} & \multicolumn{2}{c}{\bf hi} \\
 & \bf $\%$ Entities & \bf $F_1$  & \bf $\%$ Entities & \bf $F_1$ \\
\midrule
Base model & 91.8 & 67.8 & 77.9 & 37.7 \\
\hspace{8pt}+phonetic & 93.7 & 71.4 & 83.0 & 35.0 \\
\hspace{8pt}+copy & 97.2 & 72.2 & 85.4 & 37.6 \\
\hspace{8pt}+gold & 98.3 & 73.3 & 88.5 & 42.0 \\
\hspace{8pt}+dist & 99.9 & 74.2 & 94.9 & 43.4 \\
\bottomrule
\end{tabularx}
\caption{Ablation study for Spanish and Hindi}
\label{tab:ablation}
\end{table}

\paragraph{Sources of errors in TMP}
We also analyze mistakes made by TMP in aligning entities. Many false negative errors can be traced back to a high threshold $\delta$, resulting in an empty set of candidate matches. Errors also arise due to noise and variation introduced in the contextual sentence level translation of a word (Figure \ref{fig:error_analysis} (c) where GIBBS is interpreted as an acronym, (d) where MEDVEDEV is mistranslated). This causes discrepancies between translations of standalone entities and those in context, thereby, causing TMP to not find a match. However, these errors can be reduced as off-the-shelf MT systems continue to improve.

%% file: sections/discussion.tex

%% file: sections/conclusion.tex

\section{Conclusion}
In this paper, we tackled the problem of entity projection for cross-lingual NER. 
Our proposed method leverages MT for translating entities, 
matches entities based on orthographic and phonetic similarity,
and identifies matches based on distributional statistics derived from the dataset 
to achieve state-of-the-art results for cross-lingual NER on a diverse set of languages. 
Further, our method beats state-of-the-art monolingual baseline for Armenian,
an actual medium-resource language 
(off-the-shelf translation systems exist, but large-scale NER corpora do not). 
In the future, we would like to explore ways to extend our method 
to languages not supported by Google Translate through the use of pivot languages.

While dependence on MT restricts our approach to languages covered by off-the-shelf MT systems, 
these systems continue to improve in coverage and quality,
outpacing the availability of large-scale corpora for a variety of other tasks.
Moreover as translation quality improves, approaches like ours are poised to benefit.
%
%
Finally, although our method beats state-of-the-art baselines, 
not surprisingly, it falls short of NER models trained on large monolingual corpora. 
We suspect that a significant portion of this degradation is due to distribution shift 
(as evidenced by improvement in Hindi $F_1$ in Parallel regime). 
Thus one promising route to improving our models
might be to incorporate domain adaptation techniques, 
which aim to build classifiers robust to various forms of distribution shift.
